\documentclass{article}

\usepackage{arxiv}

\usepackage[utf8]{inputenc} 
\usepackage[T1]{fontenc}    
\usepackage{hyperref}       
\usepackage{url}            
\usepackage{booktabs}       
\usepackage{amsfonts}       
\usepackage{nicefrac}       
\usepackage{microtype}      
\usepackage{lipsum}		
\usepackage{graphicx}
\usepackage{natbib}
\usepackage{doi}
\usepackage{balance} 
\usepackage{amsmath}
\usepackage{algorithm}
\usepackage{algorithmic}
\usepackage{mathrsfs}
\usepackage{bm}
\usepackage{booktabs}
\usepackage{amsfonts,amssymb}
\usepackage{colortbl}
\usepackage{color}
\usepackage{tikz}
\usepackage{multicol}
\usepackage{threeparttable}
\usepackage{makecell}
\usepackage{multirow}

\title{Inferring Preferences from Demonstrations in Multi-objective Reinforcement Learning: A Dynamic Weight-based Approach}

\author{ \href{https://orcid.org/0000-0002-6014-9419}{\includegraphics[scale=0.06]{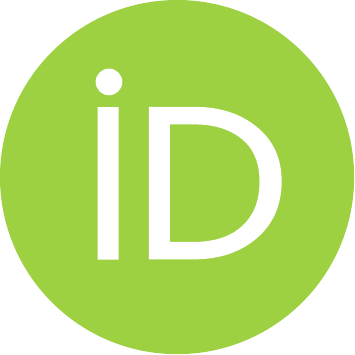}\hspace{1mm}Junlin Lu}
\\
	School of Computer Science\\
	University of Galway\\
	\texttt{J.Lu5@nuigalway.ie} \\
	\And
	\href{https://orcid.org/0000-0002-7951-878X}{\includegraphics[scale=0.06]{orcid.pdf}\hspace{1mm}Patrick Mannion} \\
	School of Computer Science\\
	University of Galway\\
	\texttt{patrick.mannion@universityofgalway.ie} \\
        \And
	\href{https://orcid.org/0000-0002-8966-9100}{\includegraphics[scale=0.06]{orcid.pdf}\hspace{1mm}Karl Mason} \\
	School of Computer Science\\
	University of Galway\\
	\texttt{karl.mason@universityofgalway.ie} \\
}

\renewcommand{\shorttitle}{Inferring Preferences from Demonstrations in Multi-objective Reinforcement Learning: A Dynamic Weight-based Approach}

\hypersetup{
pdftitle={Inferring Preferences from Demonstrations in Multi-objective Reinforcement Learning: A Dynamic Weight-based Approach},
pdfsubject={cs.AI},
pdfauthor={Junlin Lu, Patrick Mannion, Karl Mason},
pdfkeywords={First keyword, Second keyword, More},
}

\begin{document}
\maketitle

\begin{abstract}
Many decision-making problems feature multiple objectives. In such problems, it is not always possible to know the preferences of a decision-maker for different objectives. However, it is often possible to observe the behavior of decision-makers. In multi-objective decision-making, preference inference is the process of inferring the preferences of a decision-maker for different objectives. This research proposes a Dynamic Weight-based Preference Inference (DWPI) algorithm that can infer the preferences of agents acting in multi-objective decision-making problems, based on observed behavior trajectories in the environment. The proposed method is evaluated on three multi-objective Markov decision processes: Deep Sea Treasure, Traffic, and Item Gathering. The performance of the proposed DWPI approach is compared to two existing preference inference methods from the literature, and empirical results demonstrate significant improvements compared to the baseline algorithms, in terms of both time requirements and accuracy of the inferred preferences. The Dynamic Weight-based Preference Inference algorithm also maintains its performance when inferring preferences for sub-optimal behavior demonstrations. In addition to its impressive performance, the Dynamic Weight-based Preference Inference algorithm does not require any interactions during training with the agent whose preferences are inferred, all that is required is a trajectory of observed behavior. 
\end{abstract}

\keywords{Multi-objective Reinforcement Learning\and Preference Inference\and Dynamic Weight Multi-objective Agent}

\section{Introduction}
Many decision-making problems feature multiple objectives, where a trade-off between different objectives is inevitable. Numerical weights are applied to specify preferences for each objective. Varying the weights allows us to approximate the Pareto optimal set of solutions. However, it is often challenging to numerically specify a weight vector that corresponds to the user's true preference. For example, consider the multi-objective decision-making (MODM) problem of selecting stocks for a portfolio. A portfolio manager will select stocks based on their weighting of minimizing risk and maximizing potential future profits. He/she might want to give a higher weighting to maximize potential future profits, but should the weight be 0.7, 0.8, or some other number? Setting weights in this way is unintuitive and must typically be done by trial and error. Moreover, even if the user can provide an approximate numerical preference, a small error in their preference can result in a significantly different policy being learned and executed which may lead to a sub-optimal solution. 

Though it is difficult to precisely express their preference in numbers, users can always demonstrate preferences by behaviors. It would be advantageous to utilize these demonstrations to infer their preferences automatically rather than asking a user to manually give a numeric weight. Preference inference (PI) is the process of determining a user's preference over objectives and representing it numerically as, e.g., a vector in the case of linear scalarization. 

Although there is a great deal of previous work on the PI for general MODM tasks, inferring preferences from demonstrations is challenging and has received limited attention in the literature to date. Many PI works involve active learning approaches, e.g., \citet{benabbou2015incremental,zintgraf2018ordered,benabbou2020interactive}, which require feedback from users. Other approaches use ideas from inverse reinforcement learning (IRL) \citep{ng2000algorithms,ziebart2008maximum} to infer preferences from demonstrations by \citet{ikenaga2018inverse,takayama2022multi} and heuristically search for the correct weight vector in the preference space, which can be computationally expensive. It would therefore be highly advantageous to develop a new method that can perform autonomous PI at a low computational cost and without ongoing input from the user.

The central idea of our proposed dynamic weight-based preference inference (DWPI) algorithm is to train a model for PI tasks from demonstrations. The proposed algorithm is evaluated using three environments, i.e., Convex Deep Sea Treasure (CDST) \citep{mannion2017policy}, Traffic and Item Gathering \citep{kallstrom2019tunable}. In the DWPI algorithm, a dynamic weight MORL agent is trained in the environment to generate a feature set of reward trajectory and a target set of preference vectors to train a deep neural network model for policy inference. The DWPI algorithm avoids the computational overhead of heuristic approaches and requires much less user input than active learning approaches for PI. It is also more robust than other methods from the literature as it can infer preferences correctly even if the demonstrator gives a sub-optimal demonstration.

The main contributions of this paper are:
\begin{itemize}
    \item We propose the DWPI algorithm, a time-efficient method for MORL PI that utilizes demonstrations and requires no active feedback.
    \item We demonstrate the generalization ability of the DWPI algorithm by evaluating it in three environments with different characteristics, i.e., fully observable with deterministic transitions, fully observable with stochastic transitions, and partially observable with stochastic transitions.
    \item We investigate the robustness of the DWPI algorithm to sub-optimal demonstrations.
    \item We show the flexibility of the proposed DWPI algorithm by implementing it with both tabular Q Learning and Deep Q-Network.
\end{itemize} 
The remainder of this paper is organized as follows. We outline the necessary background knowledge for this paper in Section 2. In Section 3, we describe the preference inference problem. In Section 4, we present the formal model for the DWPI algorithm. In Section 5, we illustrate the design of our experiment, and the results of experiments and discuss the results. In Section 6, we conclude the paper and propose some future research directions that have arisen from this work.

\section{Background \& Related Work}
\label{sec:Background}
In this section, we review the existing works on MORL and PI in MORL.
\subsection{Multi-Objective Reinforcement Learning}
\label{subsec:Multi-Objective Decision Making}

MORL is a branch of RL that considers multi-objective decision-making problems \citep{hayes2022practical}. A MORL agent learns a policy by interacting with the environment. MORL algorithms are similar to single-objective reinforcement learning (SORL), where a reward signal is calculated to evaluate an agent action based on a state. However, in MORL, a reward vector rather than a scalar reward is given to the agent. For the convenience of training, reward scalarization is often used in MORL. The \textit{scalarization function}, also called \textit{utility function}, is the function that maps the value vector of a policy to a scalar value. Linear scalarization is one of the most frequently used scalarization methods \citep{abels2019dynamic,barrett2008learning,castelletti2012tree,roijers2017interactive}. With utility functions, we can tell whether a policy is dominated by another by comparing their utility values. A policy that maximizes the user utility can be drawn from the set of non-dominated policies. This optimal policy set is usually considered as the Pareto Optimal Set (POS) \citep{pareto2014manual}, where no one solution is said to be optimal for all objectives simultaneously.

\subsection{Preference Inference}
\label{subsec:Preference Elicitation}
Giving numerical preference vectors for MORL is not an intuitive way for a user to express their preferences \citep{tesauro1988connectionist}. The process where users provide a preference numerically is termed \textit{absolute feedback} in the work of \citep{zintgraf2018ordered}, where users' feedback based on the comparison of different solutions is noted as \textit{relative feedback}. Relative feedback is the most frequently used method in PI. This method is usually implemented as querying the user for feedback about the solutions \citep{benabbou2015incremental, zintgraf2018ordered, benabbou2020interactive}. In these works, during the process of getting feedback about the solutions, the admissible preference schemes are reduced and an approximate preference vector can be identified. This also involves frequently interacting with users during the PI process by asking questions.

\citet{ikenaga2018inverse} propose an IRL-based method is used for PI.
By comparing the expert policy's reward expectation $\mu_{E}$ and the resulting reward expectation $\mu$ using the inferred preference, the distance between ground truth preference and the inferred preference is measured. After the reward expectation from the inferred preference converges to the expectation from ground truth preference, the inference is supposed to be correct. To get the reward expectation of inferred preference, the RL agent needs to be trained from scratch every time. This is not time-efficient for high-dimensional preference spaces. To speed up the inference process, \citep{takayama2022multi} proposed an advanced method based on DRL and multiplicative weights apprenticeship learning (MWAL). Rather than updating the inference randomly in the work of \citep{ikenaga2018inverse}, they update the inference along the direction pointed out by the difference between the expert's feature expectation and the feature expectation from the inference. 

\section{Problem Statement}
\label{sec:Problem Statement}
There are similarities between concepts in PI and IRL. IRL is a methodology concerned with estimating a reward function that explains the agent's behaviour. The inferred reward function can lead to a maximum reward from the existing trajectory. In PI, assuming the utility function is the linear combination of each reward element, the aim of PI is then to infer a preference vector that maximizes the utility given the observed reward trajectory.\\
There are two assumptions for our work. The first assumption is that the behaviors/trajectories observed given are optimal or near-optimal. This is a widely accepted assumption in existing literature for both IRL \citep{ng2000algorithms,ziebart2008maximum,wulfmeier2015deep} and PI \citep{takayama2022multi}. The other assumption is that the expectation of reward trajectory is solely decided by preferences. The agent's reward trajectory is $\zeta= \{\textbf{r}_{1},\textbf{r}_{2},\dots \textbf{r}_{t}\}$. The utility of a trajectory is the sum of discounted scalarized rewards. Linear scalarization functions are the most frequently used utility function \citep{castelletti2013multiobjective,khamis2014adaptive,ferreira2017multi,lu2022multi}. Using linear scalarization function, we have Equation \ref{eqn:utility}:
\begin{equation}
    \textbf{r}_{\zeta}=\sum_{\textbf{r}_t\in{\zeta}}\gamma^{t}\textbf{r}_t
\end{equation}
\begin{equation}
\label{eqn:utility}
    utility(\textbf{r}_{\zeta}) = \bm{\omega}^{T}\textbf{r}_{\zeta}=\sum_{\textbf{r}_{t}\in{\zeta}}\bm{\omega}^{T}\gamma^{t}\textbf{r}_{t}
\end{equation}

The PI problem happens when we are given a point assumed to be on the POS based on some unknown weights for the utility function, and we would like to know what exactly the weights are. In this work, to fit a real-life scenario better, we also consider dominated points that are close to the points on the POS, known as \textit{sub-optimal policies} to extend the PI problem. In the work of \citep{ikenaga2018inverse} and \citep{takayama2022multi}, their methods do not consider sub-optimal reward trajectories where the demonstrator may not be a perfect expert for the problem. Using the heuristic method, they have a chance of making correct inferences, but their design cannot guarantee that. Even if the demonstrator cannot give a perfect demonstration that is Pareto optimal, a sub-optimal reward trajectory should be inferred to align with a preference vector close to the preference vector for the closest optimal reward trajectory.

To tackle this problem and further increase the inference speed, we trained an RL agent which can handle dynamic weights and generate a mapping between the optimal reward trajectory and preferences. This mapping function is used to train the inferring model. By adding sub-optimal noise to augment the data, we enable the inferring model to be robust on sub-optimal reward trajectories. The inferring model based on our DWPI algorithm also gives more accurate preference inference compared to \citep{ikenaga2018inverse} and \citep{takayama2022multi}.


\section{FORMAL MODEL}
\label{sec:Methodology}
In this section, we introduce the process of training a MORL agent with dynamic weights and present the proposed DWPI algorithm.

\subsection{Dynamic Weights MORL Agent Training}
\label{subsec:Train Decoder}
The DWPI algorithm is constructed based on a dynamic weights-based MORL (DWMORL) agent. We applied two algorithms to train dynamic weights MORL agents in different experiments to evaluate the robustness and generality of our DWPI algorithm.

\subsubsection{Dynamic Weights Tabular Q-learning}
The dynamic weights tabular Q-learning (DWTQ) algorithm is outlined in Algorithm \ref{alg:Q Learning with tunable training}. A DWTQ agent maintains a set of Q tables, where each of the Q tables is indexed by a possible preference vector. At the start of each episode, a preference vector is sampled from the preference space by a specific interval. The DWTQ agent interacts with the environment and updates the Q table corresponding to the preference vector until it has converged. 
\begin{algorithm}[h]\small
\caption{Dynamic Weights Multi-objective Tabular Q-learning adapted from \citep{kallstrom2019tunable}}
\label{alg:Q Learning with tunable training}
\begin{algorithmic}
\STATE {Initialize the learning rate $\alpha$, discounted factor $\gamma$}
\STATE {Initialize the number of episodes $M$, the maximum number of timesteps per episode $N$}
\STATE {Sample a list of preference vectors $\{\bm{\omega}\}$ from the preference space $\Omega$}
\FOR{each $\bm{\omega}$ \textbf{in} $\{\bm{\omega}\}$}
    \STATE {Initialize the environment}
    \STATE {Initialize a Q table $Q_{\bm{\omega}}$}
    \FOR {i\ $\xleftarrow \ 1\ \textbf{to} \  M $}
        \FOR {t\ $\xleftarrow \ 1\ \textbf{to} \  N $}
            \STATE {Take action $a_{t}$ based on state $s_{t}$ using $\epsilon-greedy$ policy}
            \STATE {Get the scalarized reward $r_t = \textbf{r}_{t}\cdot\bm{\omega}$ }
            \STATE {$Q_{\bm{\omega}}(s_t,a_t) \xleftarrow\ Q_{\bm{\omega}}(s_t,a_t) + \alpha[r_t+\gamma\ max_{a}Q_{\bm{\omega}}(s_{t+1},a)-Q_{\bm{\omega}}({s_t,a_t})]$}
        \ENDFOR
    \ENDFOR
\ENDFOR
\end{algorithmic}
\end{algorithm}

The DWTQ algorithm is proposed for simple environments with low-dimension preference spaces, i.e., Deep Sea Treasure (2-dimension preference vector). In these environments, this approach converges faster than other deep learning-based MORL dynamic weights algorithms. For more complex MO environments with higher dimension preference spaces, DWTQ would require significantly more computational time. Therefore, we propose a DQN-based PI method for higher-dimension preference spaces.  

\subsubsection{Dynamic Weights Deep Q Network}
To mitigate the scalability problem of DWTQ, the deep Q network (DQN) is used as an alternative approximator of the Q function. Unlike the traditional DQN proposed in \citep{mnih2015human}, we utilize a DQN variant \citep{kallstrom2019tunable} that can incorporate dynamic weights among objectives. 

We refer to this variant as \textit{Dynamic Weights Deep Q Network (DWDQN)}, in order to distinguish it from DWTQ. A preference vector is sampled at the start of each episode as part of the state to DWDQN. DWDQN can learn a single policy that is capable of handling preferences that change during runtime. The algorithm for training DWDQN is outlined in Algorithm \ref{alg:Tunable DQN}.
\begin{algorithm}[h]\small
\caption{Dynamic Weights Deep Q Learning, adapted from\citep{kallstrom2019tunable}}
\label{alg:Tunable DQN}
\begin{algorithmic}
\STATE {Initialize episodes number $M$, training threshold $threshold$}
\STATE {Initialize the environment, agent, replay memeory and preference space $\Omega$}
\FOR {i\ $\xleftarrow \ 1\ \textbf{to} \  M $}
    \STATE {Sample a preference vector $\bm{\omega}$ from $\Omega$}
    \STATE {done = False}
    \WHILE {not done}
            \STATE {Agent observe the state $s_{t}$ and select an action $a_{t}$ with $\epsilon$-greedy }
            \STATE {Environment returns a reward vector $\textbf{r}_{t}$ and next state $s_{t+1}$}
            \STATE{Scalarize reward $r_{t} = \textbf{r}_{t}\cdot\bm{\omega}$}
            \STATE {Put the tuple $(s_{t},a_{t},s_{t+1},r_{t},\bm{\omega})$ in the replay memory}
    \ENDWHILE
    \IF{i>$threshold$}
        \STATE {Sample batch experience from replay memory}
        \STATE {Update Q network}
    \ENDIF
\ENDFOR
\end{algorithmic}
\end{algorithm}

\subsection{DWPI Algorithm - Training Phase}
\label{subsec:Overview of the Proposed Method}
The trained dynamic weight RL agent provides a mapping function from the preference space to the reward space that conditionally depends on some optimal policy and the environment's transition. The weight vector has one entry representing the preference for each objective. During training, a preference vector is sampled from the preference space as part of the input to the Q table/DQN. This is the core of the generation of data to train the inference model. With this mapping function, we propose our DWPI algorithm to train a PI model for MODM. The inference model estimates the preferences of an expert, given the reward trajectory from a demonstration by the expert. In this paper the expert trajectories are generated by trained DWRL agents, however, the same methodology could be easily applied to infer preferences from observed expert trajectories.

The training of the inference model is conceptually the reversed process of DWRL agent training. With the hypothesis that a DWRL can always learn a policy that gives (close to) optimal utility for the preferences used during training, the process of training the preference inference model can therefore be constructed as a supervised learning problem. The feature set consists of the sum of reward vectors by the demonstrator (a DWRL agent in this paper), while the target set is the corresponding preference vector. The loss function is the mean squared error between the ground truth preference vector and the inferred preference vector. 

DWPI inferring models for the three environments use the same structure, i.e., 3 fully connected hidden layers with 64, 64, and 32 neurons each, ReLU is the activation function. The hyperparameters for training the inference model have illustrated in Table \ref{tab:Inferring Model Training Hyperparameters}. 

\begin{table}[h]\small
\centering
\caption{DWPI Inferring Model Training Hyperparameters}
\label{tab:Inferring Model Training Hyperparameters}
{\begin{tabular}{c|c|c|c}
    \toprule  \textbf{Environment}&\textbf{Loss Function} & \textbf{Optimizer}&\textbf{Learning Rate} \\
    \midrule
    \textbf{DST}&\textit{MSE}&Adam&0.001\\
    \midrule
    \textbf{Traffic}&\textit{MSE}&Adam&0.001\\
    \midrule
    \textbf{Item Gathering}&\textit{MSE}&Adam&0.005\\
    \bottomrule
\end{tabular}}
\end{table}

Another limitation of previous methods in the literature is the prerequisite that the demonstrator has provided optimal demonstrations. However, this assumption may not always hold in real life where a demonstration cannot be guaranteed as the optimal solution. DWPI addresses this limitation by introducing noise to a portion of the reward vectors to ensure that the DWPI method experiences sub-optimal demonstrations. 75\% of the reward trajectories in the PI model training dataset are sub-optimal. Adding noise to the reward trajectories ensures that the inference model does not overfit the optimal policies.

\begin{figure}[h]
    \centering
    \includegraphics[width=12cm]{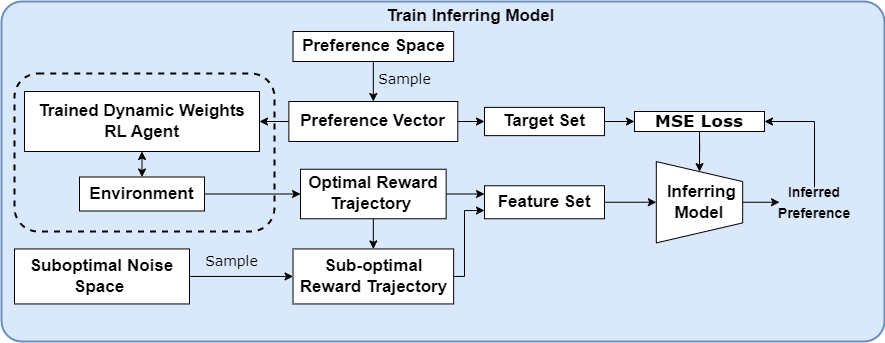}
    \caption{DWPI Training Phase}
    \label{fig:Preference Elicitator Training}
\end{figure}

As we use 3 environments that have time as an objective in the experiments in this paper, sub-optimal reward trajectories are generated by adding extra unnecessary time steps that result in an extra time penalty.
This is a realistic imitation of a true sub-optimal demonstration as a demonstrator may know how to reach their desired outcome but may not know the most efficient way to reach it. Figure \ref{fig:Preference Elicitator Training} and Algorithm \ref{alg:DWPIAlgorithm} illustrate the training phase of the DWPI algorithm. 

\begin{algorithm}[h]\small
\caption{Dynamic Weight Preference Inference Algorithm}
\label{alg:DWPIAlgorithm}
\begin{algorithmic}
\STATE {Initialize inferring model $\mathcal{I}$, sub-optimal noise space $\mathcal{SN}$, environment $\mathcal{E}$, and preference space $\Omega$}
\STATE {Load the trained dynamic weights RL agent $AG$}
\STATE {Initialize feature set $X$, target set $Y$}
\WHILE {not enough entries in $X$}
    \STATE {Sample a preference vectors $\bm{\omega}$ from $\Omega$}
    \STATE {$AG$ plays one episode with $\bm{\omega}$, generates reward trajectory $\bm{\tau_{r}}$}
    \STATE {Sample a noise vector $\bm{\delta}$ from $\mathcal{SN}$}
    \STATE {Store the noisy reward trajectory $\bm{\tau_{r}}+\bm{\delta}$ in $X$, store $\bm{\omega}$ in $Y$}
\ENDWHILE
\WHILE{$\mathcal{I}$ not converge}
\STATE {Sample batch from $X$ to train the inferring model, loss $\mathcal{L}=MSE(\hat{\bm{\omega}}, \bm{\omega})$}
\ENDWHILE
\end{algorithmic}
\end{algorithm}

\subsection{DWPI Algorithm - Evaluation Phase}
\label{subsec:DWPI Algorithm - Evaluation Phase}
The reward trajectories generated by an agent with a specific preference are firstly averaged and passed to the trained inference model for the estimation of the preference vector.

\begin{figure}[h]
    \centering
    \includegraphics[width=12cm]{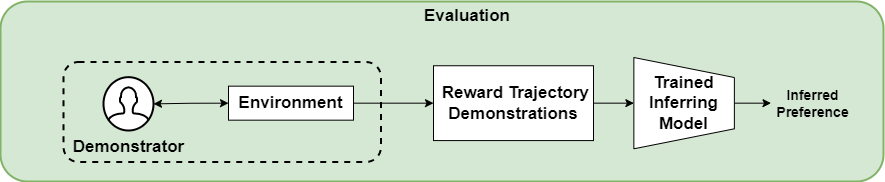}
    \caption{DWPI Evaluation Phase}
    \label{fig:Inference Phase}
\end{figure}

Figure \ref{fig:Inference Phase} outlines the process of inferring preferences from a demonstration that can be either based on optimal or sub-optimal policies.
We use multiple metrics to compare the performance of the DWPI algorithm and the baseline algorithms. Moreover, as the Pareto optimal policy is unknown in some of our experiment environments, these metrics are not used over all environments. The metrics for the environment with a known Pareto front use different metrics from the environments with an unknown Pareto front.
\subsubsection{Metric - Time Efficiency}
This metric works for all environments. Because the baseline methods \citep{ikenaga2018inverse,takayama2022multi} need to train an RL agent for each round of preference search which exponentially increases time consumption for preference spaces with higher dimensions, time efficiency is an important evaluation dimension. We compare the time consumption of DWPI and the baseline algorithms. We assigned a similar length of time budget for each of the algorithms being evaluated.
\subsubsection{Metric - Direct Comparison}
This metric works for the Convex Deep Sea Treasure \citep{mannion2017policy} environment in that the Pareto front is known. The intervals of the ground truth preference corresponding to reward trajectories are known. If the inferred preference falls within the correct interval, the inference is correct. The accuracy of the inferred result on all of the 10 possible Pareto optimal policies of CDST is calculated to evaluate the overall inference accuracy of the algorithms.
\subsubsection{Metric - Mean Squared Error}
This metric works for the Item Gathering and the Traffic environments, where typical behaviour patterns were specified in \citep{kallstrom2019tunable} but the Pareto fronts are unknown. Mean squared error is calculated between the inferred preference vector and the ground truth preference vector that specifies typical behaviour patterns. 
\subsubsection{Metric - Distributional Distance}
To check how close the distribution of the inferred preference is to the ground truth preference, KL-divergence is used to compare the results in Item Gathering and Traffic environments. 
\subsubsection{Metric - Resulting Utility}
The metrics above mathematically evaluate the performance of the algorithm. However, there are possible cases where two mathematically different preference vectors can still deliver similar utilities. If the utility from an inferred preference vector is very close to the utility from the ground truth preference, it can be regarded as a good inference. The utility from inference and the ground truth utility are compared by the magnitude of absolute errors $|u_{true}-u_{inference}|$ between the utilities.

\section{Experiment}
\label{sec:Experiment}
In this section, we outline the experimental settings and the implementation of the baseline algorithms.
\subsection{Experiment Setting}
\label{subsec:Experiment Setting}
\begin{figure}[h]
        \centering
        \includegraphics[width=12cm]{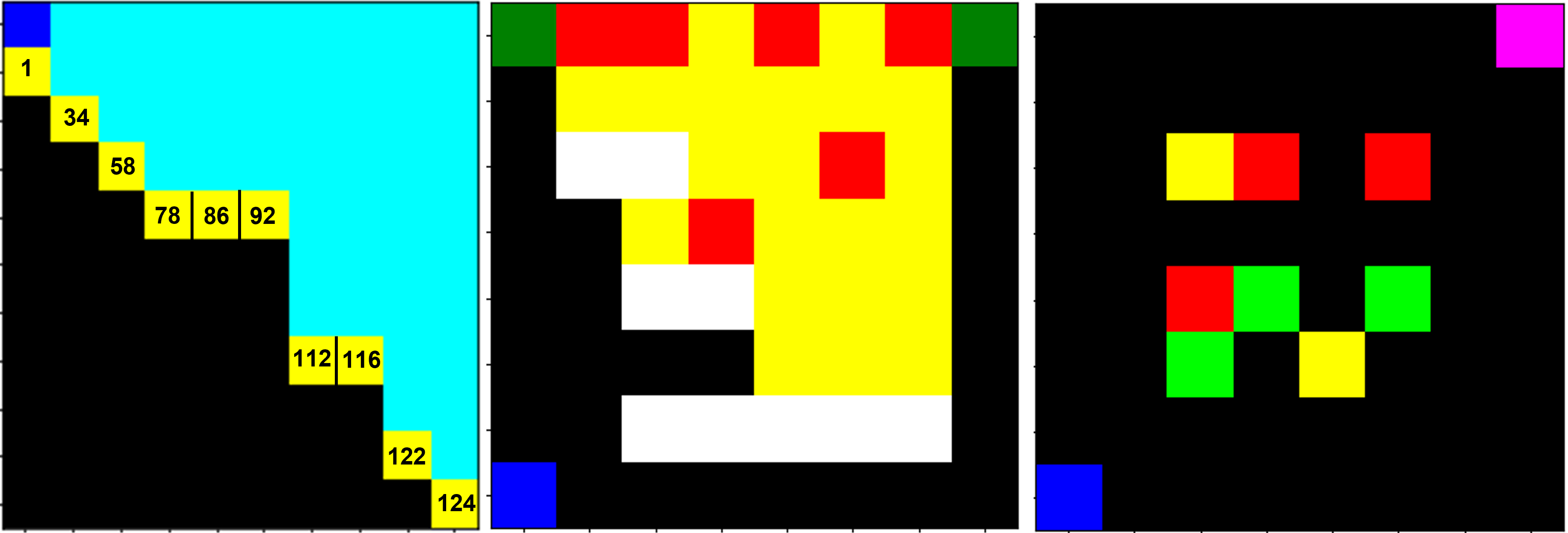}
        \caption{Convex Deep Sea Treasure Environment (left): Agent in blue, treasures in yellow with numbers, walkable grids in light blue, unwalkable grids in black. Traffic Environment (middle): Agent in blue, item to collect in green, cars in red, roads in yellow, and walls in white. Item Gathering Environment (right): Agent in blue, fixed-preference agent in pink, three categories collectable items in green, red and yellow. A fixed number of each category of items is randomly placed in the environment at the start of each episode.}
        \label{fig:DST Environment}
\end{figure}
\subsubsection{Convex Deep Sea Treasure}

The Convex Deep Sea Treasure (CDST) \citep{mannion2017policy} is a variant of Deep Sea Treasure (DST) environment \citep{vamplew2011empirical} where the globally convex Pareto front is known. The state space is the current position of the agent. The action space consists of moving up, down, left, or right. An episode starts with the agent in the left top corner and ends when the agent reaches any of the treasure grids in yellow. Each of the 10 treasures corresponds to a different policy. The agent receives a reward vector where the first element is the time penalty of -1 per time step until the episode terminates and the second element is the treasure reward it gets. The agent needs to balance between the time penalty and the treasure reward. This trade-off requires a 2-dimension preference vector over the two objectives. As the environment has a 2-dimension preference space, the DWRL agent is implemented with the DWTQ algorithm.

\subsubsection{Traffic}
The Traffic Environment \citep{kallstrom2019tunable} has a collectible item on each of the upper corners. The yellow area is the road part where the red cars move. The car moves vertically in a random direction and it reverses the direction when hit by the wall or the edge of the frame.
The agent, starting from the left bottom corner, is not supposed to step on the road. If the agent steps on the road it risks being hit by a car.
It must make a trade-off between obeying the traffic rules, maintaining traffic safety and the time it takes to get the collectible items. Each item is worth reward 1, stepping on the road causes reward -1, being hit by a car causes reward -1, hitting on the wall or the edge of the frame causes reward -1, and the time penalty is -1. In \citep{kallstrom2019tunable}, four different simulation scenarios are given, leading to four typical behavior patterns. We normalize the original preference vector in their work and fine-tune them to avoid negative weights while guaranteeing they offer the same behavior patterns. The elements of the preference vector are ordered as \textit{[steps, item collection, break traffic rules, collisions, wall hitting]}. The preference vector is rounded to 2 decimal places. The DWRL agent is implemented with DWDQN. The scenarios are shown below.\\
\textbf{Always Safe:} [0.01, 0.45, 0.09, 0.44, 0.01]\\ 
The agent tries its best to avoid illegal behavior and collisions. It takes the longest but safest path to collect the two items.\\
\textbf{Always Fast:} [0.12, 0.62, 0.12, 0.13, 0.01]\\ 
The agent tries to collect the two items as fast as possible. It thinks a detour is costly and therefore does not care about the risk of collision and breaking traffic rules by taking a short path.\\
\textbf{Fast and Safe:} [0.05, 0.47, 0.00, 0.47, 0.01]\\
The agent cares about time consumption but also the safety of the path. It is encouraged to break traffic rules and walk on the yellow path to spend less time.\\
\textbf{Slow and Safe:} [0.01, 0.49, 0.00, 0.49, 0.01]\\
The agent will walk on the road if there is a low risk of being hit by cars. It would rather take a longer path if the traffic condition is not ideal.
\subsubsection{Item Gathering}
The Item Gathering Environment \citep{kallstrom2019tunable} contains three different categories of collectibles in different colors. The positions of the items are randomly initialized at the start of each episode to evaluate the robustness of the algorithm. There is a second agent with fixed preferences that only wish to collect red items. The agent with dynamic weights, except the preferences for different categories of item and time penalty, however, also has a preference for the fixed-preference agent's success of getting a red item. This other-agent-related preference decides whether the agent behaves cooperatively or competitively. Each category item leads to reward 1, when the fixed-preference agent gets a red item this gives reward 1, and the time penalty is -1. The routes of the fixed-preference agent are randomly picked as the red items are generated randomly at the start of each episode.

Also, four different simulation scenarios are given in \citep{kallstrom2019tunable}. Original preference vectors are normalized and fine-tuned to avoid negative weights while guaranteeing they offer the same behaviour patterns. The elements of the preference vector are ordered as \textit{[steps, wall hitting, green item collected, red item collected, yellow item collected, items collected by the other agent]}. The preference vector is rounded to 2 decimal places. The DWRL agent is implemented with DWDQN. The scenarios are shown below.\\
\textbf{Competitive}: [0.02, 0.08, 0.15, 0.30, 0.15, 0.30] Cooperative Flag=0\\
The agent dislikes the other player getting the red items. It will behave competitively, and it will be harder for the fixed-preference agent to get the red items.\\
\textbf{Cooperative}: [0.02, 0.08, 0.15, 0.30, 0.15, 0.30] Cooperative Flag=1\\ 
This is the scenario where the agent would like the fixed-preference agent to get red items.\\
\textbf{Fair}: [0.01, 0.05, 0.25, 0.19, 0.25, 0.25] Cooperative Flag=1\\
The agent prefers yellow and green items to red items and it is happy with the fixed-preference agent getting red items.\\
\textbf{Generous}: [0.02, 0.08, 0.30, 0.00, 0.30, 0.30] Cooperative Flag=1\\
The agent does not care about the red items at all. It will avoid collecting red items and will be happy with the fixed-preference agent getting them.
\subsubsection{Sub-optimal Demonstration}
The sub-optimal demonstrations are given as taking 1 or 2 more unnecessary extra steps to get to the target in each of the three environments. 
\subsection{Baseline}
\label{subsec:Baseline}
We re-implement the algorithms from \citep{ikenaga2018inverse} and \citep{takayama2022multi} as the two baseline algorithms to compare our DWPI algorithm against. The algorithm by \citet{ikenaga2018inverse} is based on the apprenticeship learning and projection method (PM). Their algorithm randomly starts from a preference and compares the resulting reward trajectory with the expert trajectory. The algorithm by Takayama et al. uses multiplicative weights apprenticeship learning (MWAL). Similar to the algorithm from \citep{ikenaga2018inverse}, it infers preferences by comparing the agent's and expert's reward trajectories. Instead of randomly searching for preferences as \citep{ikenaga2018inverse}, they use Equation \ref{eqn:MWAL} to update the preference element, where $\omega_{n}$ is the nth element of the preference, k is the number of feature elements, N is the total number of iterations, and $\mu_{n}$ and $\mu_{E_{n}}$ is the nth element of the feature expectation from current RL agent and the expert.\\
\begin{equation}
\label{eqn:MWAL}
    \omega_{n}\xleftarrow{} \omega_{n}\cdot(1+\sqrt{\frac{2logk}{N}})^{-(\mu_{n}-\mu_{E_{n}})}
\end{equation}

To tailor their models to fit our environment, we change their RL part to the same algorithms, i.e. DWTQ and DWDQN algorithms.

\subsection{Results and Analysis}
\label{sec:Experiment}
For clarity, Table \ref{tab:Legend} describes the notation used in the presentation of the results. Note that for the DST environment, symbols mentioned in Table \ref{tab:Legend} all denote the treasure preference. The inference mechanism in the proposed algorithm differs from the baseline comparison methods. To ensure a fair comparison, the baseline methods terminated once the difference between the expert's rewards expectation and the expectation of rewards from the inference converges.  

\begin{figure}[h]
        \centering
        \includegraphics[width=12cm]{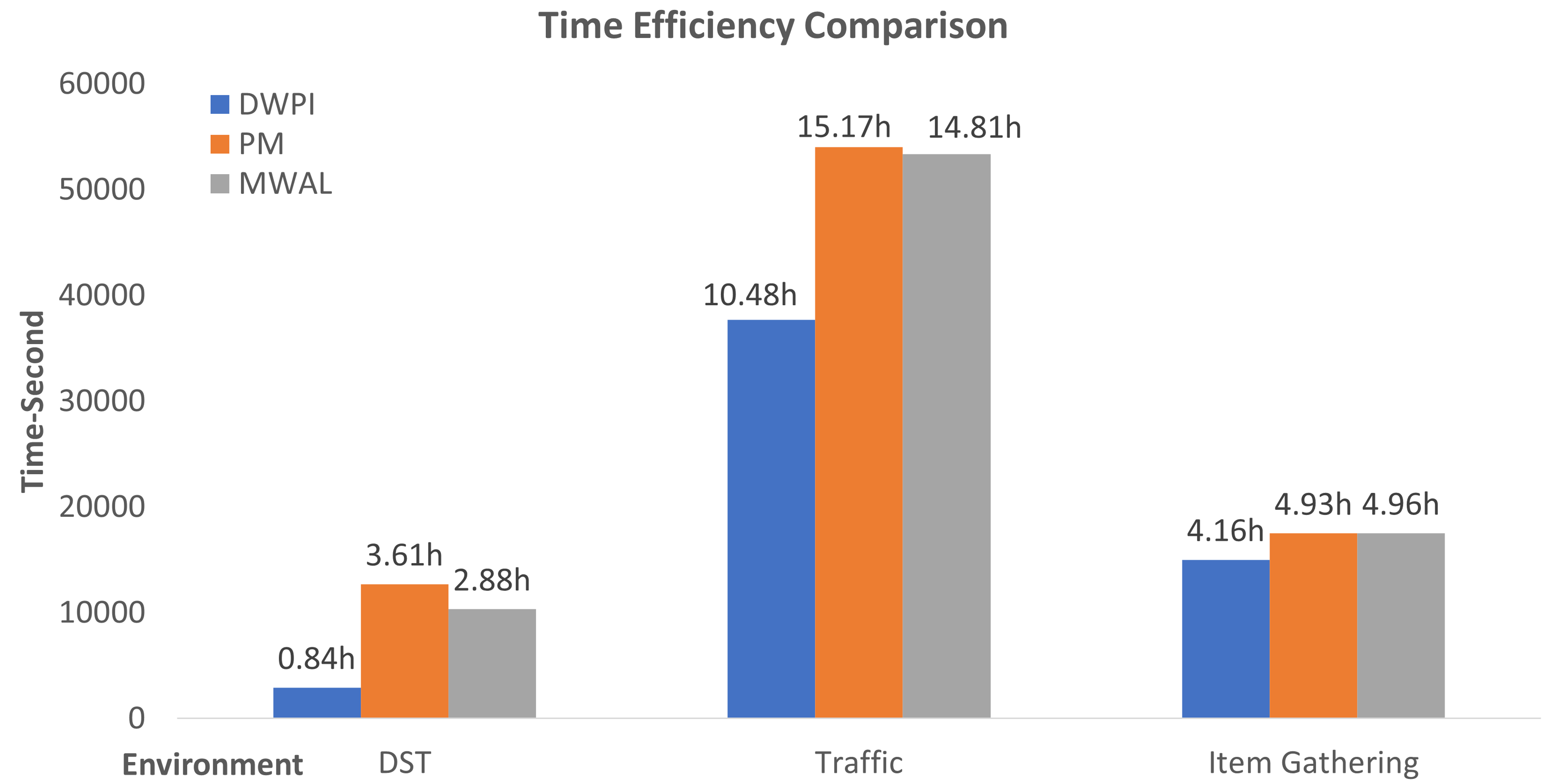}
        \caption{Time Efficiency Comparison between DWPI algorithm and Baselines.}
        \label{fig:Time Efficiency Comparison}
\end{figure}

The time requirements of each algorithm are illustrated in Figure \ref{fig:Time Efficiency Comparison}. As this figure illustrates, the DWPI algorithm outperforms the baseline PM and MWAL algorithms in terms of time requirements by 76.73\% and 70.83\% in the DST experiments, 30.92\% and 29.24\% in the Item Gathering experiments, and finally 15.62\% and 16.13\% in the Traffic experiments, respectively.

\begin{table}[h]\small
\centering
\caption{Nomenclature of Symbols}
\label{tab:Legend} 
\begin{threeparttable}
\begin{tabular}{c|c}
    \toprule
    Symbol& Definition\\
    \midrule
     $\omega$& Ground truth preference\\
     \midrule 
     $\hat{\omega}_{\pi^{*}}$& DWPI inference of optimal policy demonstrations \\
     \midrule 
     $\hat{\omega}_{\pi}$& DWPI inference of sub-optimal policy demonstrations\\
     \midrule 
     $\hat{\omega}_{BL^{1},\pi^{*}}$&PM\citep{ikenaga2018inverse} inference of optimal policy demonstrations\\
     \midrule 
     $\hat{\omega}_{BL^{1},\pi}$&PM\citep{ikenaga2018inverse} inference of sub-optimal policy demonstrations \\
     \midrule 
     $\hat{\omega}_{BL^{2},\pi^{*}}$&MWAL\citep{takayama2022multi} inference of optimal policy demonstrations\\
     \midrule 
     $\hat{\omega}_{BL^{2},\pi}$&MWAL\citep{takayama2022multi} inference of sub-optimal policy demonstrations\\
    \bottomrule
\end{tabular}
\begin{tablenotes}
\footnotesize
\item 
\end{tablenotes}
\end{threeparttable}
\end{table}

\subsubsection{DST environment}

For the DST environment, each PI algorithm is compared based on its ability to find preference values within the preference ranges that are calculated for the optimum policy. This can be done for the DST environment as the optimum policies are known for given preference ranges.

The preference vector has 2 dimensions and sums to 1. The preference for treasure objective is represented by $\omega$ while the preference for time is calculated as $1-\omega$. Therefore $\omega$ values are presented in Table \ref{tab:Preference Inference Results - DST}, where \#1 - \#10 are the 10 treasures located in DST and \textit{Acc.} is the accuracy of the inference. Incorrect inferences are colored red.
\begin{table}[h]\small
\centering
\caption{DST Preference Inference Results}

\label{tab:Preference Inference Results - DST} 
\scalebox{0.92}{
\begin{tabular}{c|c|c|c|c|c|c|c}
    \toprule
     &$\omega$&$\hat{\omega}_{\pi^{*}}$&$\hat{\omega}_{\pi}$&$\hat{\omega}_{BL^{1},\pi^{*}}$&$\hat{\omega}_{BL^{1},\pi}$&$\hat{\omega}_{BL^{2},\pi^{*}}$&$\hat{\omega}_{BL^{2},\pi}$\\
    \midrule 
     \#1&0.00-0.05&0.02 &0.02 &0   &0.03&0.04&0.03\\
    \midrule 
    \#2&0.06-0.07 &0.07 &0.06 &0.07&\textcolor{red}{0.04}&0.07&0.07\\
    \midrule 
    \#3&0.08-0.09 &0.08 &0.08 &0.09 &0.09&0.08&\textcolor{red}{0.005}\\
    \midrule 
    \#4&0.10-0.11 &0.11 &0.10 &\textcolor{red}{0.16}&0.11&0.10&0.11\\
    \midrule 
    \#5&0.12-0.14 &0.13 &0.13 &\textcolor{red}{0.18}&0.14&0.12&0.12\\
    \midrule 
    \#6&0.15-0.16 &0.15 &0.15 &0.16&0.15&\textcolor{red}{0.29}&\textcolor{red}{0.39}\\
    \midrule 
    \#7&0.17-0.20 &0.18 &0.18 &0.2&0.18&0.18&0.19\\
    \midrule 
    \#8&0.21-0.33 &0.26 &0.26 &0.21&0.24&0.31&0.22\\
    \midrule 
    \#9&0.34-0.50 &0.42 &0.40 &0.39&0.36&0.44&0.68\\
    \midrule 
    \#10&0.51-1.0 &0.75 &0.75 &0.98&0.65&0.67&0.82\\
    \midrule 
    Acc.&--&100\%&100\%&80\%&90\%&90\%&80\%\\
    \bottomrule
\end{tabular}
}
\end{table}

The results in Table \ref{tab:Preference Inference Results - DST} demonstrate that our proposed DWPI algorithm can infer the preference with 100\% accuracy when presented with both optimal and sub-optimal policies with 100\% accuracy. This is not true for either the baseline comparison algorithm.

\subsubsection{Traffic environment}
\label{subsubsec:Traffic environment}

The inference result rounded to 2 decimal points for the Traffic environment is presented in Table \ref{tab:Preference Inference Results Traffic}.

\subsubsection{Item Gathering environment}
\label{subsubsec:Item Gathering environment}
The Item Gathering environment is evaluated on the same metrics as the Traffic environment. The inference result rounded to 2 decimal points is presented in Table \ref{tab:Preference Inference Results - Item Gathering}. \textit{Cooperative Flag} (shortened as \textbf{CF}) is used to notify the DWRL agent about whether the weight for "other agent collection" is a negative value. This is done for the convenience of normalization.
\begin{table*}[h]\small
\centering
\caption{Traffic Preference Inference Results}
\label{tab:Preference Inference Results Traffic} 
\begin{tabular}{c|c|c|c|c}
    \toprule
    \textbf{Traffic} & Always Safe & Always Fast& Fast \& Safe & Slow \& Safe\\
    \midrule 
    $\bm{\omega}$ &0.01, 0.45, 0.09, 0.44, 0.01&0.12, 0.62, 0.12, 0.13, 0.01&0.05, 0.47, 0.00, 0.47, 0.01&0.01, 0.49, 0.00, 0.49, 0.01\\
    \midrule 
    $\bm{\hat{\omega}}_{\pi^{*}}$ &0.04, 0.45, 0.04, 0.44, 0.03 &0.09, 0.64, 0.10, 0.18, 0.02&0.03, 0.46, 0.01, 0.49, 0.01&0.02, 0.48, 0.01, 0.49, 0.00\\
    \midrule 
    $\bm{\hat{\omega}}_{\pi}$ &0.05, 0.44, 0.04, 0.43, 0.04&0.10, 0.60,  0.11, 0.16, 0.02&0.04, 0.46, 0.02, 0.47, 0.01&0.02, 0.48, 0.01, 0.48, 0.00\\
    \midrule 
    $\bm{\hat{\omega}}_{BL^{1}, \pi^{*}}$ &0.14, 0.71,  0.14,  0.00,  0.01 &0.07,  0.46,  0.09,  0.37,  0.01&0.09, 0.45, 0.14, 0.32, 0.00&0.09, 0.64, 0.13, 0.13, 0.01\\
    \midrule 
    $\bm{\hat{\omega}}_{BL^{1}, \pi}$ &0.10,  0.69,  0.20,  0.00,  0.01 &0.05,  0.63,  0.19,  0.13,  0.00&0.07, 0.51, 0.10, 0.31, 0.01&0.07, 0.70, 0.07, 0.14, 0.02\\
    \midrule 
    $\bm{\hat{\omega}}_{BL^{2}, \pi^{*}}$ &0.99, 0.01, 0.00, 0.00, 0.00 &1.00, 0.00, 0.00, 0.00, 0.00&1.00, 0.00, 0.00, 0.00, 0.00&1.00, 0.00, 0.00, 0.00, 0.00\\
    \midrule 
    $\bm{\hat{\omega}}_{BL^{2}, \pi}$ &0.16, 0.00, 0.00, 0.00, 0.84 &1.00, 0.00, 0.00, 0.00, 0.00&0.98, 0.00, 0.00, 0.00, 0.02&0.99, 0.01, 0.00, 0.00, 0.00 \\
    \bottomrule
\end{tabular}
\end{table*}

\begin{table*}[t]\tiny
\centering
\caption{Item Gathering Preference Inference Results}
\label{tab:Preference Inference Results - Item Gathering} 
\begin{tabular}{c|c|c|c|c}
    \toprule
     \makecell[c]{\textbf{Item}\\\textbf{Gathering}}& Competitive, \textbf{CF = 0} & Cooperative, \textbf{CF = 1}& Fair, \textbf{CF = 1} &Generous, \textbf{CF = 1}\\
    \midrule 
    $\bm{\omega}$ &0.02, 0.08, 0.15, 0.30, 0.15, 0.30&0.02, 0.08, 0.15, 0.30, 0.15, 0.30&0.01, 0.06, 0.25, 0.16, 0.26, 0.26&0.01, 0.07, 0.27, 0.11, 0.28, 0.26\\
    \midrule 
    $\bm{\hat{\omega}}_{\pi^{*}}$ &0.02, 0.08, 0.15, 0.31, 0.15, 0.30&0.01, 0.07, 0.28, 0.07, 0.28, 0.29&0.01, 0.07, 0.27, 0.09, 0.27, 0.28&0.01, 0.07, 0.28, 0.09, 0.29, 0.26\\
    \midrule 
    $\bm{\hat{\omega}}_{\pi}$ &0.01, 0.07, 0.15, 0.30, 0.15, 0.30&0.01, 0.06, 0.26, 0.16, 0.26, 0.24&0.00, 0.07, 0.24, 0.14, 0.26, 0.29&0.00, 0.07, 0.25, 0.15, 0.26, 0.28\\
    \midrule 
    $\bm{\hat{\omega}}_{BL^{1}, \pi^{*}}$ &0.02, 0.12, 0.24, 0.38, 0.12, 0.12&0.03, 0.16, 0.32, 0.00, 0.00, 0.48&0.02, 0.11, 0.11, 0.00, 0.33, 0.43&0.02, 0.12, 0.24, 0.00, 0.38, 0.24\\
    \midrule 
    $\bm{\hat{\omega}}_{BL^{1}, \pi}$ &0.03, 0.14, 0.00, 0.28, 0.42, 0.13&0.03, 0.16, 0.00, 0.32, 0.33, 0.16&0.02, 0.14, 0.00, 0.00, 0.42, 0.42&0.04, 0.16, 0.16, 0.16, 0.00, 0.48\\
    \midrule 
    $\bm{\hat{\omega}}_{BL^{2}, \pi^{*}}$ &0.99, 0.00, 0.01, 0.00, 0.00, 0.00&0.99, 0.00, 0.01, 0.00, 0.00, 0.00&0.99, 0.00, 0.01, 0.00, 0.00, 0.00&0.98, 0.00, 0.01, 0.01, 0.00, 0.00\\
    \midrule 
    $\bm{\hat{\omega}}_{BL^{2}, \pi}$ &0.03, 0.12, 0.15, 0.23, 0.36, 0.11&0.99, 0.00, 0.01, 0.00, 0.00, 0.00&0.99, 0.00, 0.00, 0.00, 0.01, 0.00&0.99, 0.00, 0.00, 0.00, 0.01, 0.00\\
    \bottomrule
\end{tabular}
\end{table*}
Due to the high-dimension nature of the Traffic and Item Gathering environments, further analysis based on the metrics discussed in Section \ref{sec:Methodology} is presented.

\begin{figure*}[h]
        \centering
        \includegraphics[width=17cm]{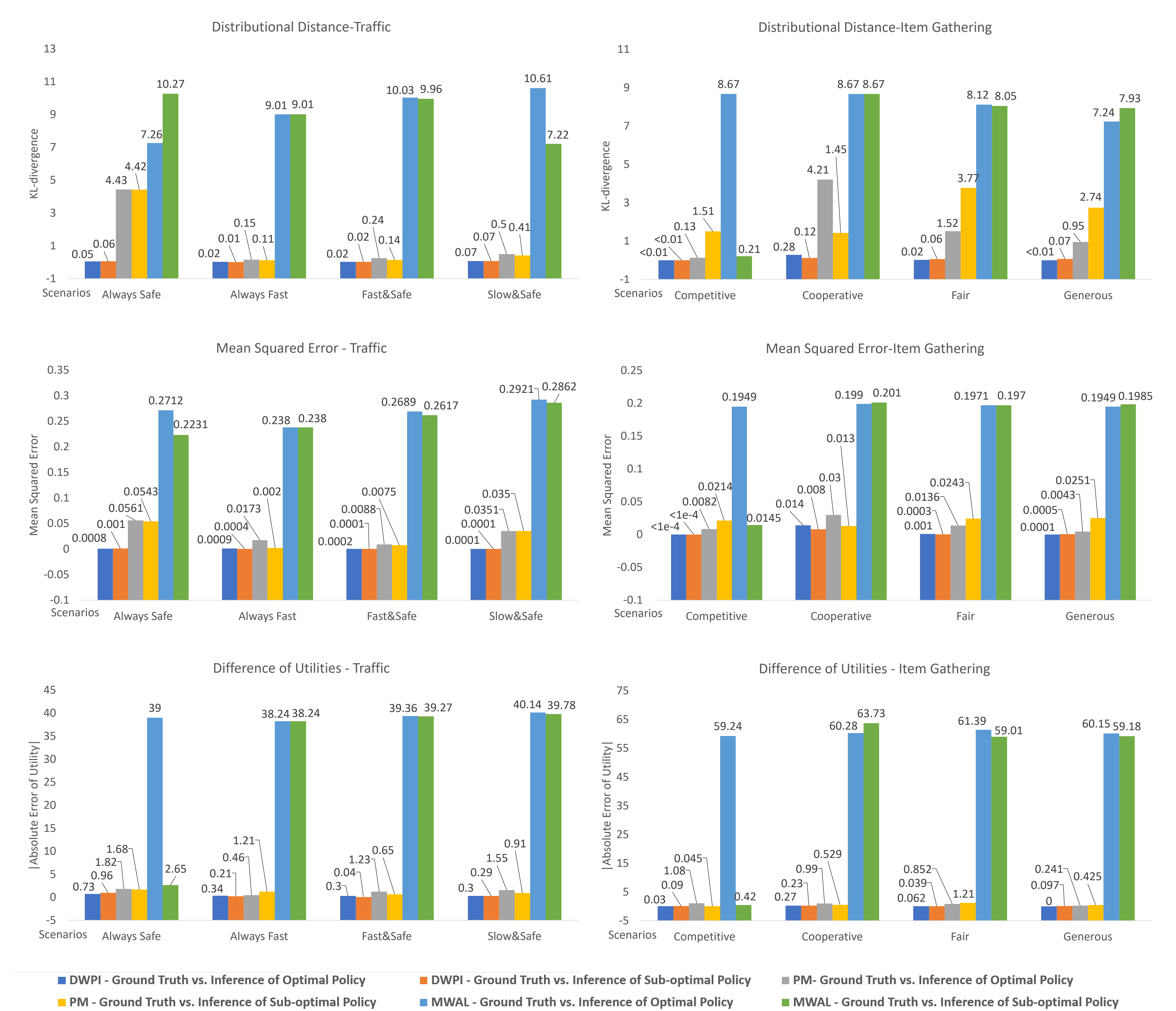}
        \caption{Analysis of Traffic and Item Gathering Results}
        \label{fig:Result of Traffic and Item Gathering}
\end{figure*}

The error metrics of each algorithm on the Traffic and Item Gathering environments are displayed in Figure \ref{fig:Result of Traffic and Item Gathering}. The distributional distance metric of the Traffic environment (upper left) and Item Gathering environment (upper right) compares the inferred preferences to the ground truth with KL-divergence between weight vectors. In case there is a preference element of 0, we replace it with an extremely small value of $10^{-5}$ to avoid computing the KL-divergence metric for values where it is not defined.

The mean squared errors of each algorithm when evaluated on the Traffic environment (Figure \ref{fig:Result of Traffic and Item Gathering} middle left) and Item Gathering environment (Figure \ref{fig:Result of Traffic and Item Gathering} middle right) are presented to illustrate the difference between the value of inference and ground truth.

To calculate the absolute error of utility (Figure \ref{fig:Result of Traffic and Item Gathering}, bottom row), a trained DWPI agent is evaluated with both the inference and ground truth on the environment by playing 100 episodes each. The average of the sum of utilities is calculated and the absolute value of the difference between the utilities is then presented. 
\subsection{Discussion}
The preference inference results demonstrate that our DWPI algorithm outperforms the baselines in all environments when evaluated for both demonstrations from optimal and sub-optimal policies on all of the five metrics. In the DST environment, the DWPI model can infer the preferences for both optimal and sub-optimal demonstrations with 100\% accuracy. This is 10pp and 20pp higher than both benchmark comparison methods for optimal and sub-optimal policies, respectively.

Table \ref{tab:Performance Improvement} summarizes the improvement in preference inference for Traffic and Item Gathering environments over the average performance for all four scenarios. In this table, the upper arrows denote the improvement over baselines. 
\begin{table}[h]\small
\centering
\caption{Performance Improvement}
\label{tab:Performance Improvement} 
\begin{tabular}{c|c|c|c|c|c}
    \toprule
    \multicolumn{2}{c}{Environment}&\multicolumn{2}{|c}{Traffic}&\multicolumn{2}{|c}{Item Gathering}\\
    \midrule
    \textbf{KL-divergence}&&PM&MWAL&PM&MWAL\\
    \midrule
    Optimal Demo&DWPI&90.8\%$\uparrow$&99.56\%$\uparrow$&98.01\%$\uparrow$&99.13\%$\uparrow$\\
    Sub-optimal Demo&DWPI&89.55\%$\uparrow$&99.53\%$\uparrow$&96.89\%$\uparrow$&99.25\%$\uparrow$\\
    \midrule
    \textbf{MSE}&&PM&MWAL&PM&MWAL\\
    \midrule
    Optimal Demo&DWPI&97.7\%$\uparrow$&99.8\%$\uparrow$&85.91\%$\uparrow$&98.1\%$\uparrow$\\
    Sub-optimal Demo&DWPI&94.13\%$\uparrow$&99.83\%$\uparrow$&83.81\%$\uparrow$&98.9\%$\uparrow$\\
    \midrule
    \textbf{Utility}&&PM&MWAL&PM&MWAL\\
    \midrule
    Optimal Demo&DWPI&60.56\%$\uparrow$&98.93\%$\uparrow$&90.67\%$\uparrow$&82.62\%$\uparrow$\\
    Sub-optimal Demo&DWPI&71.87\%$\uparrow$&90.60\%$\uparrow$&99.85\%$\uparrow$&94.50\%$\uparrow$\\
    \bottomrule
\end{tabular}
\end{table}

Apart from the improvements in terms of PI accuracy, another key benefit of the proposed DWPI over the baseline approaches is that it is significantly more computationally efficient. After training the DWRL agent and generating the set of expert trajectories and corresponding ground truth preferences, the preference inference model for DWPI can be trained in seconds for the tested domains. Furthermore, after training the inference model, it can be reused to infer preferences for any desired expert trajectory without the need for any further training. The trained DWPI preference inference model can infer preferences for any expert trajectory within a very low computational envelope (<1 second to infer the preferences for an expert trajectory for all domains tested). In contrast, the baseline approaches, PM and MWAL, are significantly less computationally efficient. With these methods, the full training process must be repeated each time a new preference is inferred for a new previously unseen expert trajectory. 

\section{Conclusion}
\label{sec:Conclusion}
This is the first study that uses a DWMORL approach for PI. The experiments conducted demonstrate that the proposed DWPI algorithm performs can infer user preferences with high accuracy. DWPI is also more computationally efficient than the existing baseline approaches from the literature. The inference model can also successfully infer preferences for sub-optimal demonstrations. 

PI for MORL is very new, and good sets of real-world benchmarks are yet to be established. Our goal is to eventually infer preference from real human demonstrations in experiments in future work.

Adding suboptimality is a promising research topic in PI. We will expand on this in the future by exploring more general ways of adding suboptimality to the demonstrator trajectory.

More avenues of future work arising from this research:\\
1. The DWPI algorithm can be evaluated in new environment types, e.g., multi-agent environments.\\ 
2. The current form of the DWPI algorithm can infer preferences for linear utility functions. In future work, we plan to extend the algorithm to be able to handle preferences expressed by non-linear utility functions \citep{agarwal2022multi}.\\
3. To apply the proposed DWPI algorithm to real-world problem domains.\\

\bibliographystyle{unsrtnat}
\bibliography{references}  






\end{document}